# On parameters transformations for emulating sparse priors using variational-Laplace inference

*J. Daunizeau*[1,2]

[1] Brain and Spine Institute, Paris, France

[2] ETH, Zurich, France

Address for correspondence:

Jean Daunizeau

Motivation, Brain and Behaviour Group

Brain and Spine Institute (ICM), INSERM UMR S975.

47, bvd de l'Hopital, 75013, Paris, France.

Tel: +33 1 57 27 43 26

Fax: +33 1 57 27 47 94

Mail: jean.daunizeau@gmail.com

Web: https://sites.google.com/site/jeandaunizeauswebsite/

Sparse priors in VBA (J. Daunizeau, 2016).

So-called *sparse estimators* arise in the context of model fitting, when one a priori assumes that only a few (unknown) model parameters deviate from zero (Li, 2007). Typically, sparsity constraints can be useful when the estimation problem is under-determined, i.e. when number of parameters to estimate ($n_\theta$) is much higher than the number of data points ($n_y$). In principle, sparsity is defined in terms of the $\ell_0$-norm, i.e. the number of non-zero parameters. However, model fitting under the constraint of minimum $\ell_0$-norm can become numerically unstable. This is why alternative approaches have been proposed, such as the so-called LASSO estimator (Tibshirani, 1996), which stands for Least Absolute Shrinkage and Selection Operator. In brief, LASSO estimators minimize the $\ell_1$-norm $\|\theta\|^1$ of model parameters $\theta$, which is simply defined as : $\|\theta\|^1 = \sum_i |\theta_i|$. LASSO estimators owe their popularity to the fact that they both produce parameters that are exactly zero and exhibit the numerical stability of $\ell_2$ regularization approaches (which do not emulate sparsity). Other alternative methods include, e.g., so-called "elastic nets", which use a mixture of $\ell_1$ and $\ell_2$ norms (Zou and Hastie, 2005), and "Horseshoe estimators", which are Bayesian estimators relying on mixture of normal priors (Carvalho et al., 2010). Note that, from a Bayesian perspective, sparsity always derives from the "fat tails" of effective priors that eventually yield the regularized estimate (Griffin and Brown, 2013).

In this note, we propose a simple parameter transform that emulates sparse priors without sacrificing the simplicity and robustness of $\ell_2$-like priors. We first show how $\ell_1$ regularization can be obtained with a "sparsify" remapping of parameters under normal priors. We then demonstrate the approach using Monte-Carlo simulations. Finally, we discuss the promising extensions to our approach.

## 1. A "sparsify" re-mapping approach to $\ell_1$ regularization

Let $\theta$ be a set of unknown parameters that shape noisy observations through some generative model $m$. For simplicity, we will consider static generative models of the form:

$$y = g(\theta) + \varepsilon \tag{1}$$

where $y$ is a set of observed data, $g(\theta)$ is an arbitrary observation mapping and $\varepsilon$ are (typically) i.i.d. Gaussian residuals.

The question we address first is how to emulate "sparse" priors for $\theta$ using well-behaved $\ell_2$-norm regularization techniques?

Recall that Bayesian approaches relying upon Gaussian priors on $\theta$ essentially yield $\ell_2$-norm regularized estimates. Our argument can be summarized as follows: one can emulate equivalent $\ell_1$-norm regularization by using Gaussian priors on transformed parameters $\tilde{\theta} = f_{sparse}(\theta)$, where the "*sparsify*" transform $f_{sparse}$ is defined as:

$$\begin{cases} f_{sparse}(\theta) = \theta^2 & \text{if } \theta \geq 0 \\ f_{sparse}(\theta) = -\theta^2 & \text{if } \theta \leq 0 \end{cases} \tag{2}$$

Critically, this mapping is monotonic, which means that ordering relationships between native and transformed parameters are preserved. In addition, this mapping is one-to-one, i.e. it uniquely maps all elements in both domain and codomain to each other.

Let us replace native parameters $\theta$ by $f_{sparse}^{-1}(\tilde{\theta})$ (where $f_{sparse}^{-1}$ is a well-defined signed square-root mapping) in the quadratic regularization term that follows from Gaussian i.i.d. priors on $\theta$:

$$p(\theta|m) = N(0,I)$$
$$\Rightarrow \log p(\theta|m) = K - \frac{1}{2}\sum_i \|\theta_i\|^2 = K - \frac{1}{2}\sum_i \|f_{sparse}^{-1}(\tilde{\theta}_i)\|^2 = K - \frac{1}{2}\sum_i |\tilde{\theta}_i| \quad (3)$$

where $K$ lumps normalization terms of multivariate normal densities.

Equation 3 means that Gaussian i.i.d. priors on native parameters $\theta$ are equivalent to a $\ell_1$-norm regularization term on transformed parameters $\tilde{\theta}$.

Now consider that, instead of using $\theta$ in the generative model (cf. Equation 1), we had used $\tilde{\theta}$, i.e.:

$$y = g(\tilde{\theta}) + \varepsilon \quad (4)$$

By definition, the LASSO estimate $\tilde{\theta}_{Lasso}$ for $\tilde{\theta}$ would optimize the following cost function:

$$\tilde{\theta}_{Lasso} = \arg\min_{\tilde{\theta}} \left[ \|y - g(\tilde{\theta})\|^2 + \lambda \|\tilde{\theta}\|^1 \right] \quad (5)$$

where $\lambda$ is the regularization parameter. If we now replace $\tilde{\theta}$ by its definition through the *sparsify* transform (Equation 2), then Equation 5 is equivalent to:

$$\theta_{Lasso} = \arg\min_{\theta} \left[ \|y - g(f_{sparse}(\theta))\|^2 + \lambda \|\theta\|^2 \right] \quad (6)$$

Equation 6 tells us that using an $\ell_2$-norm regularization term for $\theta$ (as derived from, e.g., i.i.d. Gaussian priors) and applying the *sparsify* transform on $\theta$ prior to inserting it in the observation function $g$ yields a lasso-equivalent estimate. In other words, we can happily replace Laplace priors (which would yield $\ell_1$-norm regularization terms) by gaussian priors at the cost of some non-linearity in the observation function. We can use this trick to emulate sparse priors in off-the-shelf Bayesian treatments of nonlinear models such as VBA (Daunizeau et al., 2014), which is demonstrated below.

At this point, note that the *sparsify* transform, as defined in Equation 2, is not continuous at $\theta = 0$. This may cause numerical issues during the inversion. We can deal with this by "smoothing" the *sparsify* transform as follows:

$$f_{sparse}(\theta) = (2 s_\rho(\theta) - 1)\theta^2 \tag{7}$$

where $s_\rho : \theta \rightarrow 1/1 + e^{-x/\rho}$ is the sigmoid mapping, and $\rho$ is the sigmoid temperature (which controls the amount of smoothing). At the limit $\rho \rightarrow 0$, Equation 7 recovers Equation 2 exactly, i.e. $2 s_0(\theta) - 1$ tends to the "sign" function.

Let us first inspect the impact of the *sparsify* transform, by deriving the distribution of sparsify-transformed parameters, when native parameters follow an i.i.d. Gaussian distribution. This is done by first drawing samples $\theta \sim N(0, I)$, passing these through the *sparsify* transform $f_{sparse}$, and then deriving the Monte-Carlo distribution $p(\tilde{\theta})$ of re-mapped samples $\tilde{\theta} = f_{sparse}(\theta)$.

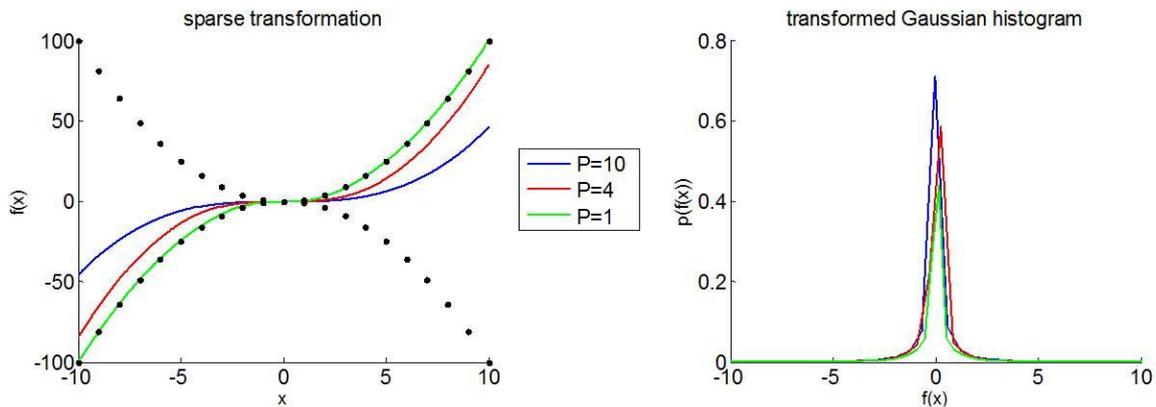

**Figure 1: induced distribution of transformed parameters under Gaussian priors on native parameters**. Left: sparsify mapping (y-axis) as a function of native parameters (x-axis), for different temperature parameters $\rho$ (colour coding). Black dots depict +/- the quadratic mapping. Right: induced probability density

function $p(\tilde{\theta})$ (y-axis) as a function of the transformed parameter values $\tilde{\theta} = f_{sparsify}(\theta)$ (x-axis), for different temperature parameters $\rho$ (same colour coding).

One can see that as the temperature decreases, the smoothed *sparsify* transform more closely matches the signed quadratic mapping. When $\rho = 1$, the distortion reduces to a very small domain of the function, and is not noticeable by the naked eye. In addition, one can see that the induced probability density function $p(\tilde{\theta})$ closely approximates a Laplace density (which was intended). What yields sparsity here is the "fat-tailed" shape of the probability density function. The effective regularization term is plotted on Figure 2 below.

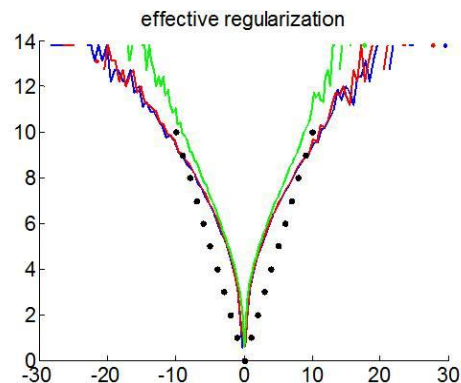

**Figure 2: effective regularization term induced by the sparsify transform (y-axis) as a function of transformed parameter values (x-axis)**. The effective regularization terms is obtained by plotting $-\log p(\tilde{\theta})$, where $p(\tilde{\theta})$ is shown on Figure 1 (right). Black dots depict the target $\ell_1$-norm regularization.

One can see that the effective regularization is sublinear, i.e. it is equivalent to a $\ell_k$-norm, where the norm order k is slightly lower than 1. This provides an intuitive justification of the sparsify transform for emulating sparse priors.

## 2. Sparsity of re-mapped Variational-Bayesian estimators

Now how does simple bayesian parameter estimates behave, when the generative model is equipped with Gaussian priors, and the observation function is distorted using the sparsify transform?

Without loss of generality, let us consider the case of Gaussian priors on $\theta$ (i.e. $p(\theta|m) = N(0, \alpha^2 I)$) and i.i.d. Gaussian residuals (i.e.: $p(\varepsilon|m) = N(0, \sigma^2 I)$), where $m$ is the generative model. Recall that the so-called *Variational Laplace* or VL approach (Daunizeau et al., 2009; Friston et al., 2007) uses a fixed-form Gaussian approximation $q(\theta)$ to the posterior density $p(\theta|y, m)$, i.e.: $q(\theta) \triangleq N(\mu, \Sigma)$ is assumed to be Gaussian with first two moments $\mu$ and $\Sigma$ given by:

$$\begin{cases} \mu = \arg\max_{\theta} I(\theta) \\ \Sigma = -\left[ \frac{\partial^2 I}{\partial \theta^2} \bigg|_{\mu} \right]^{-1} \end{cases} \quad (8)$$

where the variational energy $I(\theta)$ is itself given by:

$$I(\theta) = -\frac{1}{2}\left( n_y \ln 2\pi + \ln \sigma^2 + \frac{1}{\sigma^2}(y - g(\theta))^T (y - g(\theta)) + n_\theta \ln 2\pi + \ln \alpha^2 + \frac{1}{\alpha^2} \theta^T \theta \right) \quad (9)$$

At convergence, the VL approach also yields an approximation $F \approx \log p(y|m)$ to the (log-)model evidence:

$$F = I(\mu) + \frac{1}{2}\ln|\Sigma| + \frac{n_\theta}{2}\ln 2\pi \quad (10)$$

As we will see below, the approximate posterior mean $\mu$ can serve as a VL parameter estimate, whereas the above "free energy" $F$ can be used for model selection purposes. In Equations 8-10, $n_y$ (resp. $n_\theta$) denotes the number of observations (resp. parameters) and $g(\theta)$ is the arbitrary observation function of Equation 1.

One can see (cf. Equation 9) how the variational energy resembles a typical $\ell_2$-regularization cost function, where the regularization term follows from Gaussian priors on model parameters. In particular, it is easy to show that the relative weight of the regularization term is directly controlled by the ratio $\sigma^2/\alpha^2$. Note that the VL variant we use below is slightly more sophisticated than the one we describe here because it also estimates the variance hyperparameter $\sigma^2$. This effectively allows VL to adapt the amount of regularization to one's specific data descriptive statistics. In what follows, we refer to this VL variant as VBA (Daunizeau et al., 2014). However, this also means that we expect the amount of regularization to be dependent upon the data signal-to-noise ratio...

Following the first section of this draft, $\ell_1$-like sparsity is emulated by first re-mapping the model parameters through the *sparsify* transform, i.e. the observation function now becomes: $g(\theta) \to g \circ f_{sparse}(\theta)$. Note that this does not change the VL learning rule (Equation 8), given that it was defined for any arbitrary observation mapping.

Finally, one may be interested in inferring the posterior probability $P(\theta = 0 | y)$ that some parameter is zero. This can be derived using a Bayesian model comparison between the original (full) model $m$ and a reduced (null) model $m_0$ that assumes *a priori* that $P(\theta = 0 | m_0) = 1$ (e.g., using a Dirac prior probability mass at $\theta = 0$):

$$P(\theta = 0 | y) = \frac{p(y | m_0)}{p(y | m_0) + p(y | m)} \approx \frac{1}{1 + \exp(F - F_0)} \quad (11)$$

where $F_0$ is the free energy under model $m_0$. Practically speaking, $P(\theta = 0 | y)$ can be derived using so-called Savage-Dickey ratios without having to invert the reduced model (Marin and Robert, 2010; Penny and Ridgway, 2013).

Sparse priors in VBA (J. Daunizeau, 2016).

Let us now evaluate the sparsity properties of VL estimators. We do this by simulating data under a specific case of generative models considered here, namely: general linear models or GLMs. We do this in such a way that the problem of estimating the parameters is ill-posed, simply because the number of unknown parameters ($n_\theta = 128$) is twice the number of data observations ($n_y = 64$). The native observation mapping in Equation 1 now simply reduces to $g(\theta) = X\theta$, where $X$ is an arbitrary $n_y \times n_\theta$ design matrix. In what follows, we set $X$ by drawing i.i.d. random gaussian samples.

First, let's simulate data with sparse parameters, i.e. where only a few simulated parameters are non-zero ("sparse simulations"). Figure 3 below summarizes the output of VBA, when we invert the model under either Gaussian priors, or under sparse priors (using the sparsify transform trick).

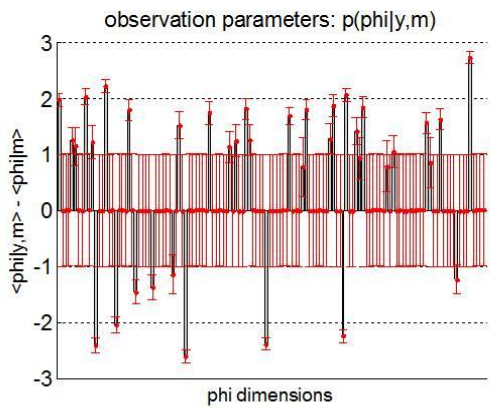
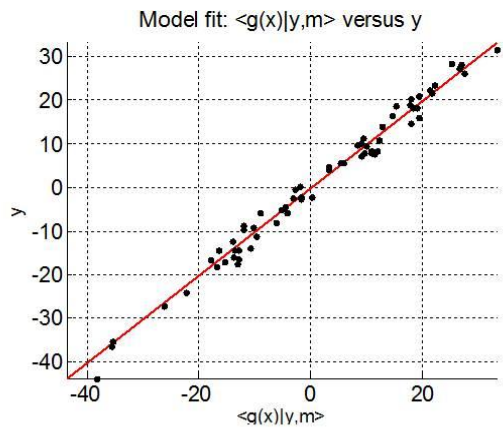
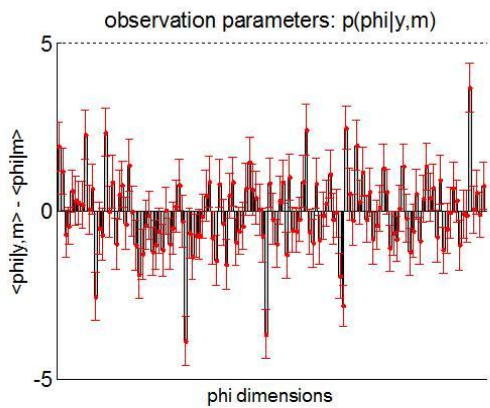
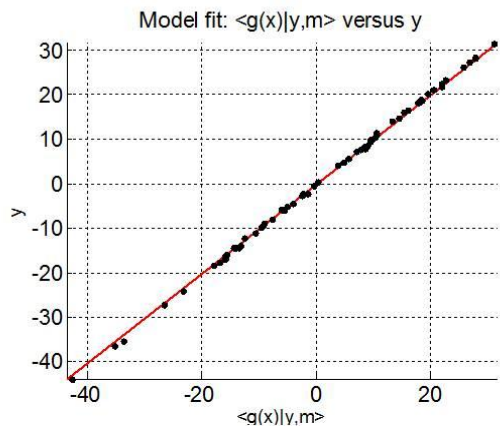

**Figure 3: "Sparse simulations": posterior density (left) and model fit (right) under sparse priors (upper panels) and under Gaussian priors (lower panels).** Note that in this case, model comparison (where models differ in terms of whether one includes or not the sparsify transform into the GLM) favours the sparse priors: ΔF = 25.1.

As a control condition, let us simulate data with non-sparse parameters, i.e. where simulated parameters are small but non-zero ("Gaussian simulations"). Figure 4 below summarizes the output of VBA, when we invert the model under both types of priors.

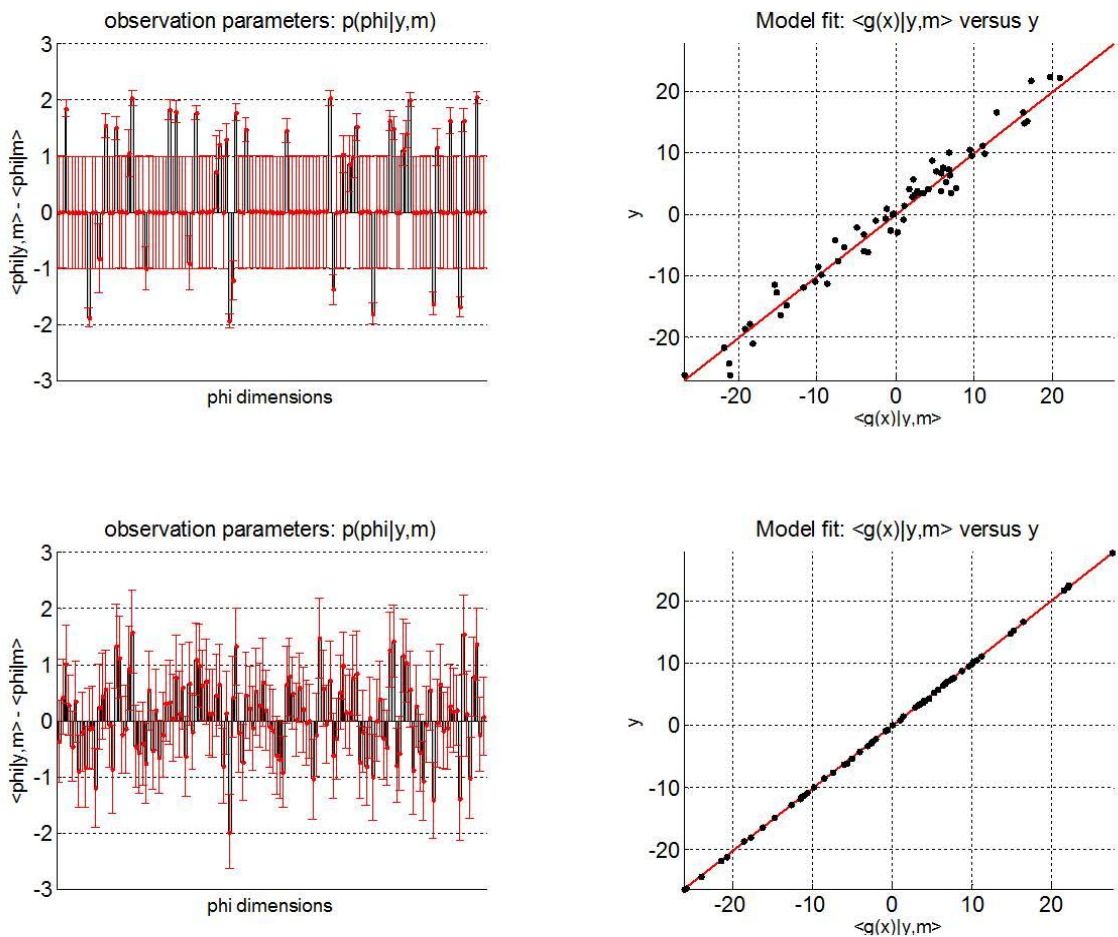

**Figure 4: "Gaussian simulations": posterior density (left) and model fit (right) under sparse priors (upper panels) and under Gaussian priors (lower panels).** Here, model comparison favours the Gaussian priors: ΔF = -16.2.

Sparse priors in VBA (J. Daunizeau, 2016).

One can see the impact of the sparsify transform on parameter estimation in both types of simulations. In brief, most parameters are back to their prior specification (posterior=prior), except for a handful of these. This eventually yields much bigger residuals than when using Gaussian priors, i.e. sparse priors protect from overfitting (Reunanen, 2003). In addition, Bayesian model comparison identifies the correct type of priors, in that, for both types of simulations, the parameter estimates under the winning model are the most accurate. This can be checked by measuring, e.g., the correlation between simulated and estimated parameters (cf. Figure 5 below).

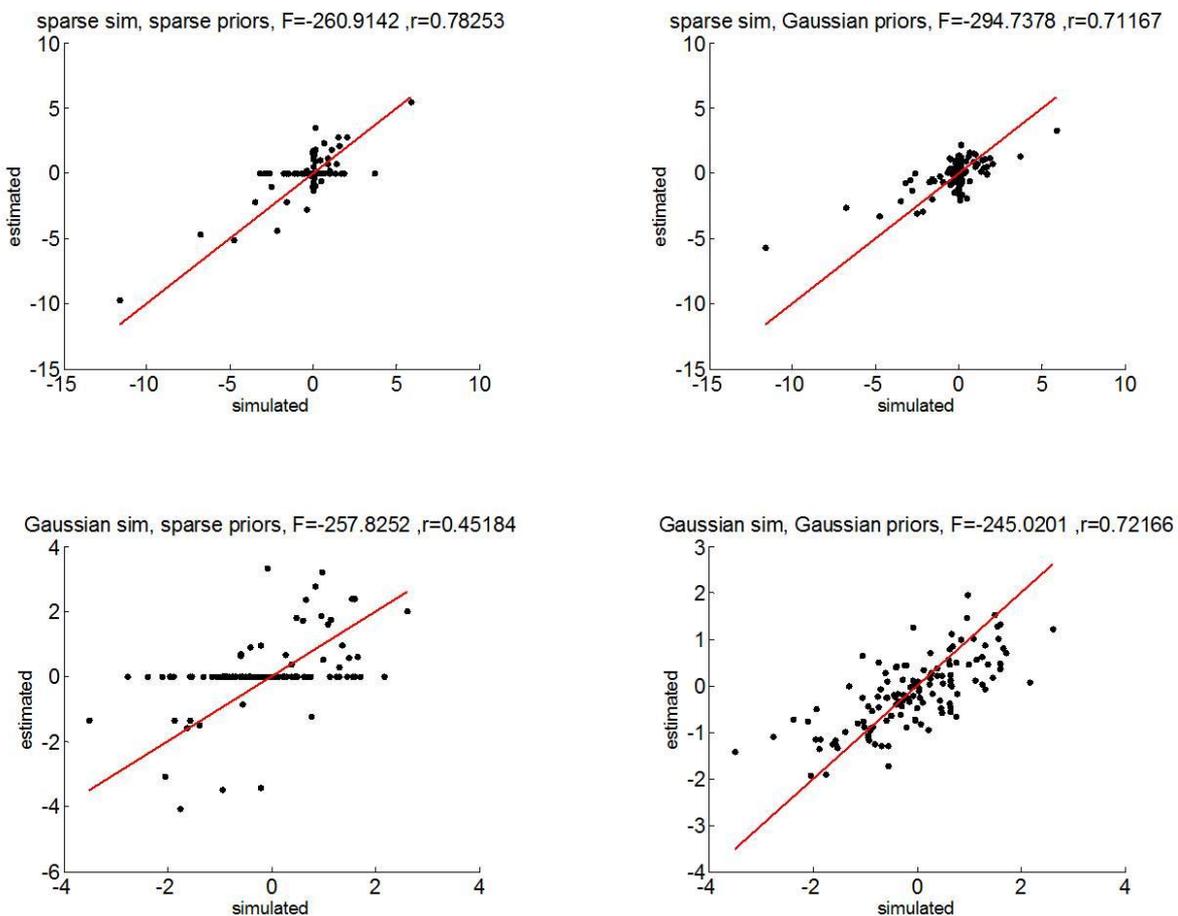

**Figure 5: Parameter estimation accuracy for both sparse (upper panels) and gaussian (lower panels) simulations under sparse (left) and gaussian priors (right).** Estimated parameters (y-axis) are plotted against simulated parameters (x-axis) for each type of simulations and priors. The red line depicts perfect match. The sample correlation is indicated above each graph (along with the corresponding free energy).

We then performed Monte-Carlo simulations to evaluate the respective impacts of the level of sparsity and SNR onto both estimation accuracy and VL's ability to recognize the best estimator (sparse versus gaussian). We systematically varied the error precision $\sigma^{-1}$ (from 0.01 to 100) and the simulated sparsity (in terms of the rate of zero parameters). For each possible pair of data precision and sparsity, we simulated 128 datasets, by randomly sampling the model parameters as well as measurement noise. Both "sparse" and "gaussian" models were then inverted given each simulated dataset. We then quantified the evidence in favour of the "sparse" model (in terms of $F_{sparse} - F_{gaussian}$) and the difference in estimation accuracy (in terms of the difference in correlations between simulated and estimated parameters $r_{sparse} - r_{gaussian}$). Figure 5 below summarizes the results of these Monte-Carlo simulations.

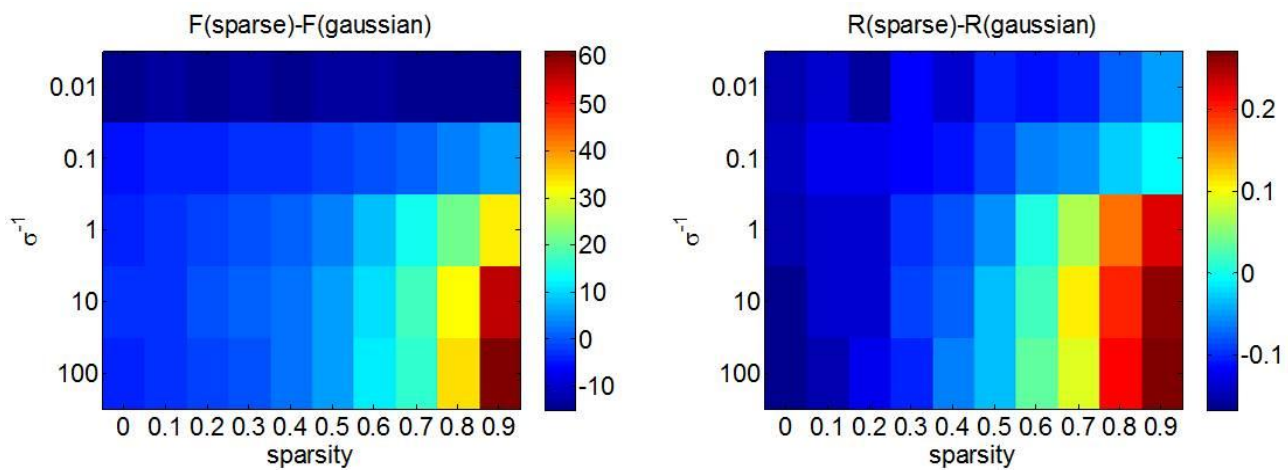

**Figure 5: Evidence in favour of "sparse" model (left) and added-value of "sparse" estimation accuracy (right), as a function of simulated sparsity (x-axis) and data precision (y-axis).** Each element in these images is the average over 128 Monte-Carlo simulations.

One can see that both metrics follow the same pattern, namely they increase with simulated sparsity and data precision. Recall that one would expect that both metrics would increase with simulated sparsity. The effect of data precision, however, is less trivial. We will discuss this effect later.

We then looked directly at the average relationship between both metrics, which is summarized in Figure 6 below.

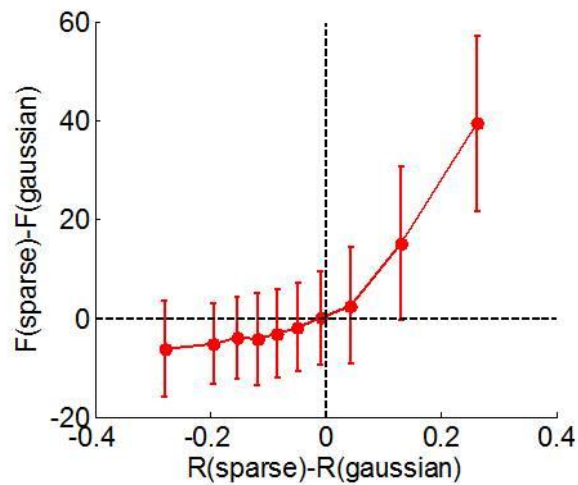

**Figure 6: Evidence in favour of "sparse" model (y-axis) as a function of added-value of "sparse" estimation accuracy (x-axis)**. The difference in correlations was first binned in ten quantiles, and the mean and standard deviation (over Monte-Carlo simulations) of the corresponding difference in free energies was then derived.

One can see that there is a good adequacy between the evidence in favour of the "sparse" model and the difference in estimation accuracies. In particular, there is no bias in the relative amount of evidence, i.e. $F_{sparse} - F_{gaussian}$ becomes positive only when $r_{sparse} - r_{gaussian}$ is itself positive. This means that, on average, Bayesian model comparison actually selects the model that achieves the highest estimation accuracy.

We also asked whether the domains over which parameters are zero or not are accurately recovered. In other terms, we asked whether the VL sparse estimator is effectively zeroing the right model parameters. Figure 7 summarizes the Monte-Carlo average of both true positive (TPR) and true negative rates (TNR). Both TPR and TNR are derived from thresholding the posterior probability $P(\theta = 0|y)$. Here, we used a frequentist-like approach and thresholded this probability to a target (uncorrected) FPR of 5%.

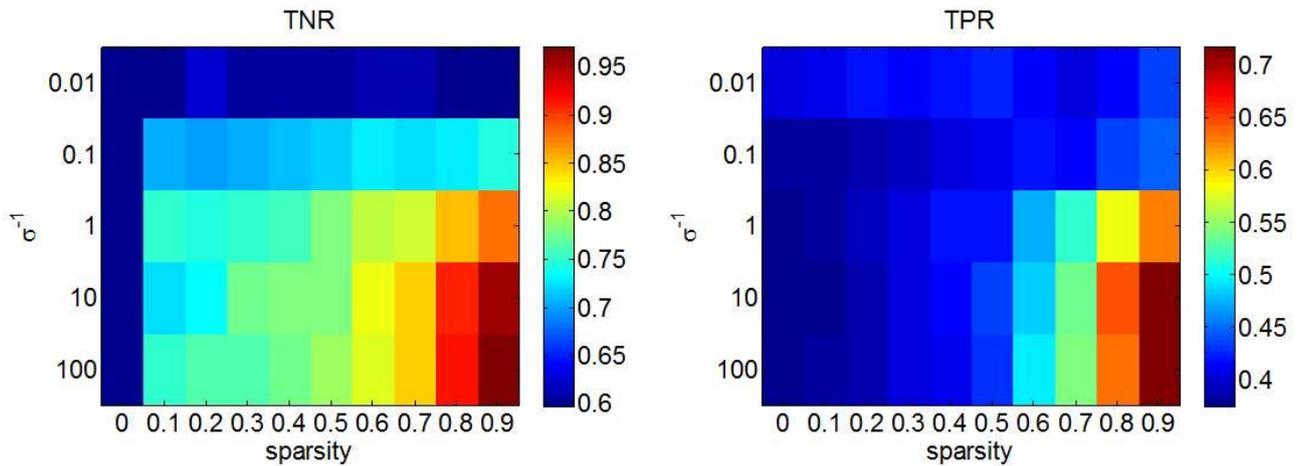

**Figure 7: TNR (left) and TPR (right), as a function of simulated sparsity (x-axis) and data precision (y-axis)**. Each element in these images is the average over 128 Monte-Carlo simulations.

One can see that, here again, both TNR and TPR increase with simulated sparsity and data precision. In particular, VL sparse estimation is decently discriminative, since both TPR and TNR are, most of the time, above chance level (despite the under-determination of the problem, cf. $n_y = n_\theta/2$). Recall that, here, one would expect that both TPR and TNR increase with data precision. This is simply because $\sigma^{-1}$ directly controls the quality of the information that can be extracted from the data. But no simple prediction of this sort could have been made *a priori* regarding the effect of simulated sparsity. As we will see, this effect can be best understood when looking at the estimated sparsity, which directly derives from the thresholded posterior probability $P(\theta = 0|y)$. This is summarized in Figure 8 below.

Sparse priors in VBA (J. Daunizeau, 2016).

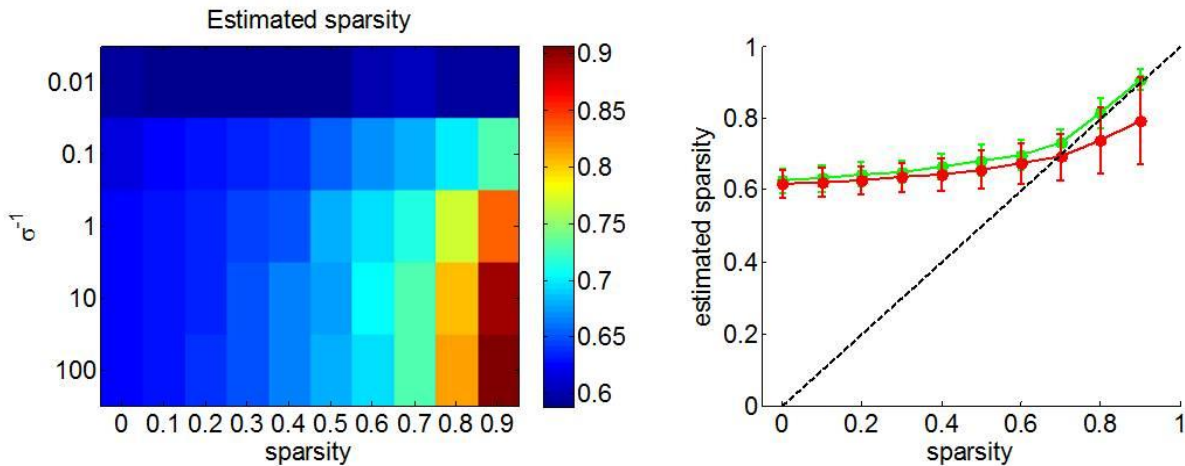

**Figure 8: Estimated sparsity.** Left: estimated sparsity is depicted as a function of simulated sparsity (x-axis) and data precision (y-axis). Each element in this image is the average over 128 Monte-Carlo simulations. Right: estimated sparsity (y-axis, red: averaged over data precision levels, green: for $\sigma^{-1}=100$) is plotted as a function of simulated sparsity (x-axis) only.

One can see that the estimated sparsity increases with both data precision and simulated sparsity. In particular, there is a monotonic and positive relationship between estimated and simulated sparsity. However, one can see that moderately sparse simulations are incorrectly recovered, in that the minimal sparsity that is achieved (about 60%) does not match simulated sparsity. In other words, VBL tends to overestimate sparsity. This means that VL-sparse estimators are conservative, in the classical (frequentist) sense: they tend to neglect some effects that are, strictly speaking, non-zero.

There are two reasons for this. The first one is that some simulated parameters where trivially small. Recall that, when performing Monte-Carlo simulations, we draw simulated parameters from a normal density with zero mean and unit variance. Thus, some non-zero simulated parameters may have a minor impact on the data, and may eventually be deemed negligible.

The second reason relates to data informativeness. In brief, we expect the estimated sparsity to more closely match the simulated sparsity when the problem becomes better conditioned, e.g. when the ratio $n_y/n_\theta$ increases. To illustrate this point, we performed the same series of Monte-Carlo simulations, but this time with $n_y = n_\theta = 32$. Qualitatively speaking, this does not change (i) the statistical relationship between the relative evidence in favour of the sparse model and the added-value of sparse estimation, and (ii) the impact of data quality and simulated sparsity upon TPR and TNR. In more quantitative terms however, estimation accuracy, TPR, TNR and sparsity estimation improve. To illustrate this, we summarize the results in terms of Figure 9 below, which reproduces the analysis of Figure 8 (with $n_y = n_\theta = 32$).

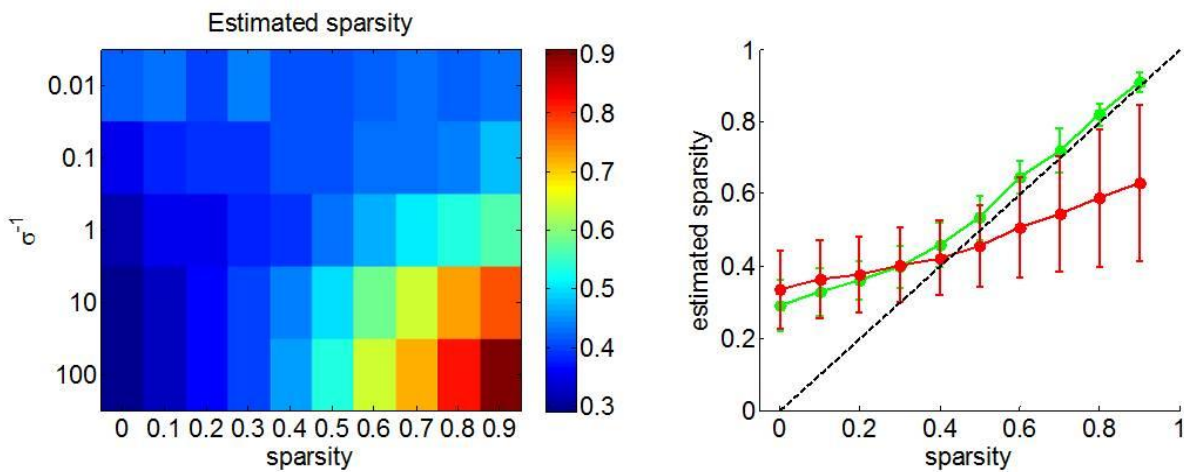

**Figure 8: Estimated sparsity for $n_y = n_\theta = 32$.** Same format as in Figure 8.

One can see that sparsity is better estimated when the problem becomes better conditioned. In particular, if the magnitude of model residuals is reasonably small (e.g., when $\sigma^{-1} = 100$), then estimated sparsity becomes very accurate.

## 3. Discussion

In conclusion, we have proposed a simple parameter transform that emulates sparse priors without sacrificing the simplicity and robustness of $\ell_2$-like priors. We have shown how $\ell_1$ regularization can be obtained with a "sparsify" remapping of parameters under normal priors, and we have demonstrated the ensuing variational Bayesian (VB) approach using Monte-Carlo simulations.

Our numerical investigation of sparse-VL estimation has identified two interesting properties. First, we have shown that Bayesian model selection correctly predicts which regularization scheme ($\ell_1$ versus $\ell_2$) eventually yields most accurate parameter estimates. This is important, since this provides robustness to the ensuing parameter estimate. Second, sparse-VL estimation seems to be slightly conservative, and this tendency decreases with data quality and quantity. From a classical (frequentist) perspective, this is acceptable, in contrast to liberal approaches that would tend to exhibit an overly elevated false alarm rate.

Note that all these simulations can be retrieved from the script `demo_sparsePriors.m` from the VBA freeware (https://mbb-team.github.io/VBA-toolbox/).

It is well known that estimating the domain of non-zero parameters and minimizing overall estimation error are two different problems that are best addressed with different regularization constraints, such as those enforced using $\ell_1$ or $\ell_2$ norms, respectively (Wang et al., 2014). Thus, one may want to design adaptive sparse estimators that adjust the effective norm of the regularization term. This calls for parametric forms of the sparsify transform, e.g.:

$$f_{sparse}(\theta) = (2\,s(\theta) - 1)|\theta|^{\omega} \tag{12}$$

where $\omega$ controls the effective order of the norm. For example, if $\omega = 1$, then the sparsify mapping is (almost) linear and the ensuing VL approach yields $\ell_2$-regularized estimates. If $\omega = 2$, then VL yields $\ell_1$-regularized estimates. If $1 \leq \omega \leq 2$, then VL estimates are equipped with intermediary levels of sparsity. In addition, at the limit $\omega \to \infty$, VL estimates are derived under a constraint of minimum $\ell_0$-norm. Critically, within a Bayesian approach, the hyperparameter $\omega$ could be included in the generative model, and estimated along with model parameters $\theta$. This constitutes a potentially interesting extension of this work, which we will investigate in forthcoming publications.

# References

Carvalho, C.M., Polson, N.G., and Scott, J.G. (2010). The horseshoe estimator for sparse signals. Biometrika *97*, 465–480.

Daunizeau, J., Friston, K.J., and Kiebel, S.J. (2009). Variational Bayesian identification and prediction of stochastic nonlinear dynamic causal models. Phys. Nonlinear Phenom. *238*, 2089–2118.

Daunizeau, J., Adam, V., and Rigoux, L. (2014). VBA: A Probabilistic Treatment of Nonlinear Models for Neurobiological and Behavioural Data. PLoS Comput Biol *10*, e1003441.

Friston, K., Mattout, J., Trujillo-Barreto, N., Ashburner, J., and Penny, W. (2007). Variational free energy and the Laplace approximation. NeuroImage *34*, 220–234.

Griffin, J.E., and Brown, P.J. (2013). Some Priors for Sparse Regression Modelling. Bayesian Anal. *8*, 691–702.

Li, L. (2007). Sparse Sufficient Dimension Reduction. Biometrika *94*, 603–613.

Marin, J.-M., and Robert, C.P. (2010). On resolving the Savage–Dickey paradox. Electron. J. Stat. *4*, 643–654.

Penny, W.D., and Ridgway, G.R. (2013). Efficient Posterior Probability Mapping Using Savage-Dickey Ratios. PLOS ONE *8*, e59655.

Reunanen, J. (2003). Overfitting in Making Comparisons Between Variable Selection Methods. J. Mach. Learn. Res. *3*, 1371–1382.

Tibshirani, R. (1996). Regression Shrinkage and Selection via the Lasso. J. R. Stat. Soc. Ser. B Methodol. *58*, 267–288.

Wang, J., Lu, C., Wang, M., Li, P., Yan, S., and Hu, X. (2014). Robust Face Recognition via Adaptive Sparse Representation. IEEE Trans. Cybern. *44*, 2368–2378.

Zou, H., and Hastie, T. (2005). Regularization and variable selection via the elastic net. J. R. Stat. Soc. Ser. B Stat. Methodol. *67*, 301–320.